\documentclass[9pt,conference]{IEEEtran}
\IEEEoverridecommandlockouts

\usepackage{cite}
\usepackage{amsmath,amssymb,amsfonts,enumitem}
\usepackage{dsfont}
\usepackage{algorithmic}
\usepackage{graphicx}
\usepackage{textcomp}
\usepackage{xcolor}
\usepackage{multicol,multirow}
\def\BibTeX{{\rm B\kern-.05em{\sc i\kern-.025em b}\kern-.08em
    T\kern-.1667em\lower.7ex\hbox{E}\kern-.125emX}}

\usepackage{rotating}
\definecolor{KleinBlue}{HTML}{8a2be2}
\definecolor{OliveGreen}{rgb}{0,0.6,0}
\definecolor{shadecolor}{rgb}{0.8,0.8,0.8}
\definecolor{DarkRed}{rgb}{0.55, 0.0, 0.0}

\usepackage{xstring}
\usepackage{ifthen}

\newcommand{\diff}[1]{%
  \if\relax\detokenize{#1}\relax %
    \textcolor{DarkRed}{#1}%
  \else
    \IfBeginWith{#1}{+}{%
      (\textcolor{OliveGreen}{#1})%
    }{%
      (\textcolor{DarkRed}{#1})%
    }%
  \fi
}

\begin{document}

\title{Controllable Forgetting Mechanism for Few-Shot Class-Incremental Learning}

\author{\IEEEauthorblockN{Kirill Paramonov\IEEEauthorrefmark{1}, Mete Ozay\IEEEauthorrefmark{1}, Eunju Yang\IEEEauthorrefmark{2}, Jijoong Moon\IEEEauthorrefmark{2}, Umberto Michieli\IEEEauthorrefmark{1}\thanks{\textcopyright 2025 IEEE.  Personal use of this material is permitted. Permission from IEEE must be obtained for all other uses, in any current or future media, including reprinting/republishing this material for advertising or promotional purposes, creating new collective works, for resale or redistribution to servers or lists, or reuse of any copyrighted component of this work in other works.}}
\IEEEauthorblockA{\IEEEauthorrefmark{1}Samsung R\&D Institute UK (SRUK), Staines, Surrey, United Kingdom}
\IEEEauthorrefmark{2}Samsung Research, Seoul R\&D Campus, Seoul, Rep. of Korea\\
Email: \{k.paramonov, m.ozay, ej.yang, jijoong.moon, u.michieli\}@samsung.com}
\maketitle

\begin{abstract}
Class-incremental learning in the context of limited personal labeled samples (few-shot) is critical for numerous real-world applications, such as smart home devices. A key challenge in these scenarios is balancing the trade-off between adapting to new, personalized classes and maintaining the performance of the model on the original, base classes. Fine-tuning the model on novel classes often leads to the phenomenon of catastrophic forgetting, where the accuracy of base classes declines unpredictably and significantly.
In this paper, we propose a simple yet effective mechanism to address this challenge by controlling the trade-off between novel and base class accuracy. We specifically target the ultra-low-shot scenario, where only a single example is available per novel class. Our approach introduces a Novel Class Detection (NCD) rule, which adjusts the degree of forgetting \textit{a priori} while simultaneously enhancing performance on novel classes.
We demonstrate the versatility of our solution by applying it to state-of-the-art Few-Shot Class-Incremental Learning (FSCIL) methods, showing consistent improvements across different settings. To better quantify the trade-off between novel and base class performance, we introduce new metrics: NCR@2FOR and NCR@5FOR. Our approach achieves up to a 30\% improvement in novel class accuracy on the CIFAR100 dataset (1-shot, 1 novel class) while maintaining a controlled base class forgetting rate of 2\%.
\end{abstract}


\section{Introduction}
In recent years, deep learning models have become integral to many mobile devices and home appliances for computer vision tasks \cite{zhang2023deep,min2023large}. For instance, a food recognition application on a mobile device can be pre-trained on a fixed set of dishes (e.g., Western cuisine) and deployed on the device \cite{raghavan2024online,heng2021cnfoot}. Such an application can accurately recognize a dish if it belongs to one of the pre-trained classes. However, when a user requests recognition of an unseen dish (e.g., an Asian dish that was not part of the pre-training), the model will likely fail. To address this, it is essential to incorporate continual learning capabilities, allowing the system to learn from a few user-provided images of the novel dish and recognize future instances of that dish \cite{raghavan2024online,he2023long,nguyen2023incremental,luo2023class}.

\begin{figure}[htbp]
    \centerline{
        \includegraphics[trim=0cm 12.8cm 21.7cm 0cm, clip,width=1\linewidth]{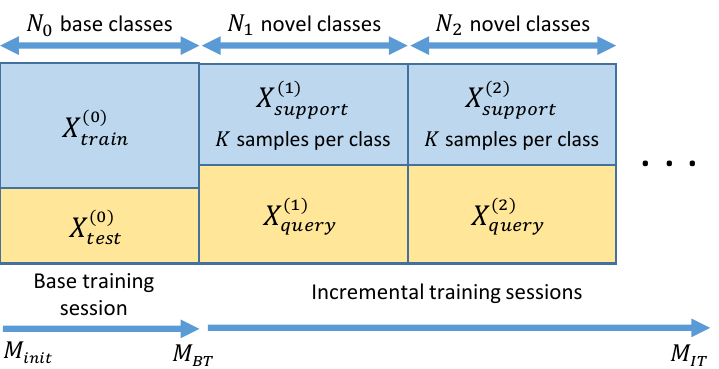}
    }
    \caption{Setup for FSCIL with $K$ shots. A \textit{base training} session is usually done on the server and multiple \textit{incremental training} sessions are usually done on device with a few annotated samples (i.e., the support set) from novel classes. In our paper, we focus on the one-shot case ($K=1$).}
    \label{fig:setup}
\end{figure}

This setup, known as Few-Shot Class-Incremental Learning (FSCIL) \cite{tao2020few,tian2024survey,zhang2021few,wang2021few}, involves two key stages: a \textit{base training} session and several \textit{incremental training} sessions (see Fig~\ref{fig:setup}). In the base session, the model is trained on a large number of samples from \textbf{base} classes (e.g., Western dishes). Once deployed, the model encounters a few annotated examples from \textbf{novel} classes (e.g., Asian dishes) and must adapt to improve its accuracy on these new classes while still retaining knowledge of the base classes.

Most FSCIL solutions consider incremental sessions that introduce 5–10 novel classes, each with 5 labeled samples (or \textbf{shots}) \cite{cheraghian2021semantic,song2023learning,zhou2022forward,shi2021overcoming}. However, this setting is often unrealistic in real-world applications, where users may be unwilling to provide multiple annotated samples for each new class. In this paper, we tackle a more challenging scenario: One-Shot Class-Incremental Learning (OSCIL), where each incremental session consists of only a single annotated sample per novel class.

\begin{figure*}[tbp]
    \centering
    \includegraphics[trim=0cm 11.4cm 7.6cm 0cm, clip,width=1\linewidth]{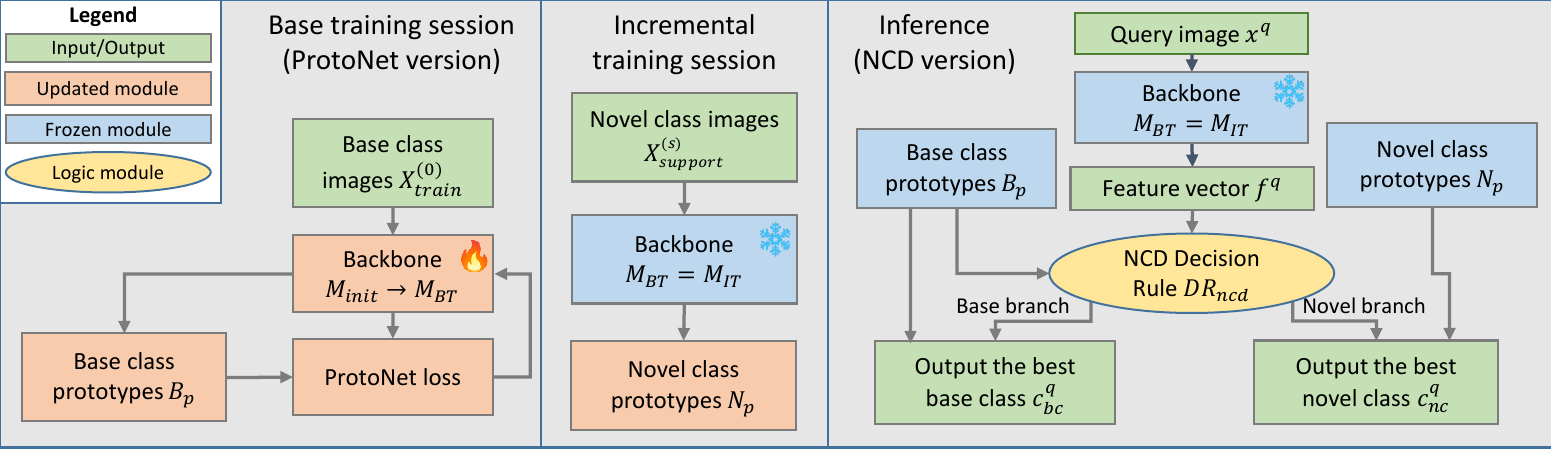}
\caption{Overview of our method. Left: base training session (Sec~\ref{sec:base_train}), e.g., based on ProtoNet~\cite{snell2017prototypical}, SAVC~\cite{song2023learning}, FACT~\cite{zhou2022forward}, OrCo~\cite{ahmed2024OrCo}. Middle: incremental training session (Sec~\ref{sec:incremental_train}) with frozen backbone. Right: inference stage (Sec~\ref{sec:inference}), where our NCD \textbf{Decision Rule} controls the inference logic flow between branches of base and novel classes.}
\label{fig:method}
\end{figure*}

Another significant challenge in FSCIL is preserving the accuracy of base class recognition during incremental updates. Fine-tuning the model on new classes typically boosts novel class accuracy but leads to a pronounced drop in base class performance, a phenomenon known as \textbf{catastrophic forgetting} \cite{tao2020few,cheraghian2021semantic,dong2021few}. This problem is exacerbated in low-shot settings and on low-resource devices, where retaining base class samples on the device is impractical, and a single novel sample is insufficient to fine-tune the network via backpropagation effectively.
To mitigate catastrophic forgetting, existing works focus on two main strategies: improving either the base training session \cite{song2023learning,zhou2022forward,shi2021overcoming,paramonov2024swiss,michieli2024object,michieli2023online} or the incremental training sessions \cite{wu2024generalizable,cheraghian2021semantic,dong2021few,tao2020few,zhao2021mgsvf}. The former aims to develop a backbone that performs well on base classes and generalizes to novel ones, with the backbone typically frozen during incremental sessions. The latter approach introduces small additional modules during base training, which are selectively updated in the incremental sessions.

In this paper, we present the following contributions:
\setlist{nolistsep}
\begin{enumerate}[itemsep=0em]
    \item We propose a novel inference method for OSCIL based on a branching decision rule that significantly enhances novel class recognition accuracy while controlling the trade-off between base and novel class performance.
    \item We introduce \textbf{controllable forgetting}, allowing predictable and adjustable base class forgetting during adaptation to novel classes, tailored for low-resource devices without the need to store old samples.
    \item Our approach is plug-and-play compatible with existing state-of-the-art base training methods.
    \item As a by-product, our method facilitates \textbf{out-of-distribution (OOD) detection} for query images (i.e., determining whether an image belongs to previously seen base classes). This capability is valuable for real-world applications, enabling the system to autonomously prompt the user to annotate novel images.
\end{enumerate}

\section{Methodology}

In this section, we outline our setup (see Fig.~\ref{fig:setup}) and introduce our proposed Novel Class Detection (NCD) method for OSCIL with controllable forgetting (see Fig.~\ref{fig:method}).

\subsection{Few-Shot Class-Incremental Recognition}

FSCIL typically begins with an initial backbone model $M_{init}$, which is often pre-trained on larger datasets (e.g., ImageNet-1k) using either a cross-entropy loss (e.g., ResNet) or a self-supervised contrastive loss (e.g., DINO Transformer).

Next, the model undergoes a \textit{base training} phase on domain-specific data, focusing on a fixed set of \textbf{base classes} with abundant samples ($\gg 100$ per class). The output of this phase is the domain-specific model $M_{BT}$, trained on the base classes.
We denote the base class train split as $X^{(0)}_{train}$, test split as $X^{(0)}_{test}$, and the number of base classes as $N_0$.

Once trained, $M_{BT}$ is deployed on a personal device, where it encounters a few annotated samples from previously unseen \textbf{novel classes} (e.g., Asian food dishes). This leads to continual stream of \textit{incremental training} sessions, where the model adapts to novel class data and produces $M_{IT}$. 
For an incremental training session $s>0$, we denote annotated (or \textbf{support}) samples from novel classes as $X^{(s)}_{support}$, test (or \textbf{query}) samples from novel classes as $X^{(s)}_{query}$, the number of novel classes as $N_s$ and the number of support samples (or \textbf{shots}) per class as $K$.
Without loss of generality, we focus on a case with a single incremental session $s=1$, since we can combine all samples we've seen so far into a single support set $X^{(1)}_{support}$, and perform incremental training session on it.
To account for that, we provide results for varying number of classes in the first session $N_1$.

The main challenge in FSCIL is to balance between adapting to novel classes and retaining knowledge about the base classes during the incremental training stage.
Overly fitting the model to the novel classes leads to forgetting of the base classes.
Specifically, we focus on three key metrics:
\begin{itemize}[itemsep=0em]
    \item \textbf{Base class recognition:} accuracy on the test split of base classes after the base training session.
    \vspace{-0.8em}
    \begin{equation}
        \mathrm{BCR} := \mathrm{ACC}(M_{BT}; X^{(0)}_{test}).
        \label{eq:bcr}
        \vspace{-0.8em}
    \end{equation}
    \item \textbf{Novel class recognition:} accuracy on the novel class query samples after incremental training.
    \vspace{-0.8em}
    \begin{equation}
        \mathrm{NCR} := \mathrm{ACC}(M_{IT}; X^{(1)}_{query}).
        \label{eq:ncr}
        \vspace{-0.8em}
    \end{equation}
    \item \textbf{Base class forgetting rate:} decline in base class accuracy due to learning new classes.
    \vspace{-0.8em}
    \begin{equation}
        \mathrm{FOR} := \mathrm{BCR} - \mathrm{ACC}(M_{IT}; X^{(0)}_{test}).
        \label{eq:for}
        \vspace{-0.5em}
    \end{equation}
\end{itemize}

In this paper, we target the challenging one-shot setting, where each novel class has only a single annotated sample—a scenario closer to real-world applications but less explored in literature that currently focuses on 5 or 10 shots.

\subsection{Base Training Session}
\label{sec:base_train}

Our method is focused on the inference stage (see Fig.~\ref{fig:method}) and is agnostic to the choice of base training procedure. 
To evaluate effectiveness of our inference method, we apply it on top of two base training procedures: the popular ProtoNet training~\cite{snell2017prototypical} and the state-of-the-art SAVC~\cite{song2023learning}, OrCo~\cite{ahmed2024OrCo} and FACT~\cite{zhou2022forward}.
Both methods, as well as most current FSCIL methods, rely on the notion of \textbf{prototypes}, which denotes a centroid of the class-wise feature vectors. 
For a given class $c$, the prototype is defined as
\vspace{-0.5em}
\begin{equation}
    \mathrm{proto}_c := \mathrm{Avg}_i(M(x_c^i)),
    \label{eq:proto}
    \vspace{-0.5em}
\end{equation}
where $M$ is a backbone model, and $x_c^i\in X_{train}^{(0)}$ is the $i$-th annotated sample of base class $c$.

ProtoNet employs prototypical loss which is more effective and robust for few-shot learning applications. SAVC and FACT use contrastive learning and augmented base classes in the base training session to effectively partition the feature space. OrCo promotes orthogonality between features.

After the base training session, we store the set of base class prototypes (denoted as $\mathcal{B}_p$) in memory and use it during inference.

\subsection{Incremental Training Session}
\label{sec:incremental_train}

Following prior FSCIL approaches~\cite{song2023learning, zhou2022forward, shi2021overcoming}, we freeze the backbone during incremental training to prevent \textit{uncontrollable} forgetting. 
During this phase, we compute and store the prototypes from Eq.~\ref{eq:proto} for novel classes ($\mathcal{N}_p$) from their support samples in $X_{support}^{(1)}$.

\subsection{Decision Rule for Inference Stage}
\label{sec:inference}

During inference in standard (vanilla) FSCIL methods, a query sample $x^q$ is assigned to the class whose prototype is closest in feature space \cite{mensink2013distance}:
\vspace{-0.5em}
\begin{equation}
    c_{pred, van}^q = \mathrm{argmin}_c\big(\mathrm{dist}(f^q, \mathrm{proto}_c)\big),
    \label{eq:vanilla}
    \vspace{-0.3em}
\end{equation}
where $f^q := M(x^q)$ is the feature vector of a query sample $x^q$, $\mathrm{proto}_c$ are stored prototypes from base and novel classes ($\mathcal{B}_p \cup \mathcal{N}_p$), and $\mathrm{dist}(\cdot)$ is a distance function in the feature space (typically cosine distance).

However, in the one-shot setting, the variability in novel class support samples introduces noise into the prototype estimation, leading to inaccurate predictions. 
To alleviate that issue, we introduce a boolean function \textbf{Decision Rule} $DR(f^q)$ to control the logic flow for the inference. 
In general, a decision rule is based on both base class prototypes $\mathcal{B}_p$ and novel class prototypes $\mathcal{N}_p$ (i.e., $DR(f^q)=DR(f^q; \mathcal{B}_p, \mathcal{N}_p)$), and can have different designs.
In this paper, we design a specific decision rule for the one-shot task, which we call \textbf{Novel Class Detection (NCD)}. NCD reduces the dependency of noisy support samples on the final inference, relying more on the stable base class centroids.
In particular, NCD assigns a sample to a novel class if its feature vector is far away from all base prototypes:
\vspace{-0.8em}
\begin{equation}
    DR_{ncd}(f^q; \alpha, \mathcal{B}_p) := \mathds{1}\big(\min_{\mathrm{proto_c} \in \mathcal{B}_p} \mathrm{dist}(f^q, \mathrm{proto_c}) > \alpha \big),
    \vspace{-0.5em}
\end{equation}
where $\mathds{1}$ is the indicator function, and $\alpha$ is a pre-defined distance threshold.

The resulting predicted class is then:
\begin{equation}
    c_{pred,ncd}^{q,\alpha} := 
    \begin{cases}
        c_{nc}^q,  & \text{if } DR_{ncd}(f^q; \alpha, \mathcal{B}_p) \text{ is True},\\
        c_{bc}^q,  & \text{otherwise},
    \end{cases}
    \label{eq:dr}
\end{equation}
where $c_{bc}^q$ corresponds to the closest prototype from $\mathcal{B}_p$ and $c_{nc}^q$ corresponds to the closest prototype from $\mathcal{N}_p$.

\subsection{Controllable Forgetting}

Denoting $M_{IT, ncd}^\alpha$ the model with our NCD rule and distance threshold $\alpha$, note that we can calculate the accuracy of the model on base class samples without knowledge of novel class samples, since the NCD rule does not take the personal samples into account.

Therefore, we can calculate the accuracy $\mathrm{ACC}(M_{IT, ncd}^\alpha; X^{(0)}_{test})$ a-priori before deploying the model on device.
So, we can \textbf{control} the forgetting rate $\mathrm{FOR}$ from Eq.~\eqref{eq:for} by setting appropriate distance threshold $\alpha$ based on the pre-defined forgetting budget for the base classes.
We call this feature of our method \textbf{controllable forgetting}, in which the base class recognition accuracy will always be within the forgetting budget, regardless of the encountered novel samples.
This feature is crucial for on-device applications, where maintaining a predictable level of base class accuracy is essential for ensuring the quality of service.

\begin{table*}[htbp]
    \caption{Main results for $N_1=1$ and $N_1=5$ and two levels of pre-defined forgetting rate (2 and 5, respectively). Our strategies (\textit{NCR@2FOR} and \textit{NCR@5FOR}) consistently outperforms vanilla inference strategy, especially in the 1-shot regime. Relative gains are show compared to \textit{V-NCR}.}
    \begin{center}
    \renewcommand{\arraystretch}{1.3}
    \begin{tabular}{|c|c|c|c|c|c|c|c|c|c|}
        \hline
        \multirow{2}{*}{\textbf{Backbone}} & \textbf{Parameter} & \multirow{2}{*}{\textbf{Dataset}} &
        \textbf{Base} & \multicolumn{3}{|c|}{$\mathbf{N_1=1}$} & \multicolumn{3}{|c|}{$\mathbf{N_1=5}$} \\
        \cline{4-10} 
        & \textbf{Count (M)} & & \textit{BCR} & \textit{V-NCR} & \textit{NCR@2FOR} & \textit{NCR@5FOR} & \textit{V-NCR} & \textit{NCR@2FOR} & \textit{NCR@5FOR} \\
        \hline
        \textbf{MobileNetv2-PN} & 3.5 & \textbf{CUB200} & 77.2 & 19.2 & {30.0} \diff{+10.8} & {51.1} \diff{+31.9} & 14.7 & {18.6} \diff{+3.9} & {30.8} \diff{+16.1} \\
        \hline
        \textbf{MobileNetv2-SAVC} & 3.5 & \textbf{CUB200} & 69.2 & 34.6 & {32.5} \diff{-2.1} & {56.9} \diff{+22.3} & 27.0 & {21.8} \diff{-5.2} & {35.3} \diff{+8.3} \\
        \hline
        \multirow{2}{*}{\textbf{ResNet18-PN}} & \multirow{2}{*}{11.6} & \textbf{CUB200} & 76.8 & 14.9 & {24.2} \diff{+9.3} & {41.9} \diff{+27.0} & 18.3 & 16.1 \diff{-2.2} & {29.4} \diff{+11.1} \\
        \cline{3-10} 
        & & \textbf{CIFAR100} & 76.6 & 11.3 & 16.3 \diff{+5.0} & 35.5 \diff{+24.2} & 13.1 & 15.9 \diff{+2.8} & 32.5 \diff{+19.2} \\
        \hline
        \multirow{2}{*}{\textbf{ResNet18-SAVC}} & \multirow{2}{*}{11.6} & \textbf{CUB200} & 74.5 & 25.3 & {26.3} \diff{+1.0} & {47.6} \diff{+22.3} & 22.8 & 19.8 \diff{-3.0} & {36.3} \diff{+13.5} \\
        \cline{3-10} 
        & & \textbf{CIFAR100} & 78.4 & 16.5 & 22.5 \diff{+6.0} & 43.8 \diff{+27.3} & {13.6} & 17.1 \diff{+3.5} & 30.4 \diff{+16.8} \\
        \hline
        
        \multirow{2}{*}{\textbf{ResNet18-OrCo}} & \multirow{2}{*}{11.6} & \textbf{CUB200} & 71.6 & 12.6 & 27.0 \diff{+14.4} & 42.8 \diff{+30.2} & 15.0 & 21.4 \diff{+6.4} & 35.5 \diff{+20.5} \\
        \cline{3-10} 
        & & \textbf{CIFAR100} & 79.0 & 17.2 & 26.2 \diff{+9.0} & 39.4 \diff{+22.2} & 10.1 & 12.7 \diff{+2.6} & 22.5 \diff{+12.4} \\
        \hline
        \multirow{2}{*}{\textbf{ResNet18-FACT}} & \multirow{2}{*}{11.6} & \textbf{CUB200} & 71.5 & 27.3 & 38.8 \diff{+1.5} & 61.2 \diff{+33.9} & 23.6 & 23.7 \diff{+0.1} & 42.4 \diff{+18.8} \\
        \cline{3-10} 
        & & \textbf{CIFAR100} & 84.7 & 17.3 & 29.3 \diff{+12.0} & 48.3 \diff{+31.0} & 17.0 & 21.0 \diff{+4.0} & 35.8 \diff{+18.8} \\
        \hline
        
        \textbf{DINOv2s-init} & 22.0 & \textbf{CORe50} & 73.0 & 55.8 & 57.5 \diff{+1.7} & 77.8 \diff{+22.0} & 58.6 & {59.8} \diff{+1.2} & {61.1} \diff{+2.5} \\
        \hline
        \textbf{DINOv2s-PN} & 22.0
        & \textbf{CORe50} & 85.4 & 60.7 & 79.4 \diff{+18.7} & 86.9 \diff{+26.2} & 61.3 & 55.2 \diff{-6.1} & 60.6 \diff{-0.7} \\
        \hline
    \end{tabular}
    \label{tab:main}
    \end{center}
    \vspace{-10pt}
\end{table*}

\section{Experiments}

\subsection{Backbone Models and Datasets}

\textbf{Backbone Models.} 
We evaluate the effectiveness of our Novel Class Detection (NCD) rule using three different backbone architectures with varying complexity to account for different resource constraints during deployment: \textbf{MobileNetV2}\cite{sandler2018mobilenetv2}, \textbf{ResNet18}\cite{he2016deep}, and \textbf{DINOv2s}~\cite{oquabdinov2}. 
During the base training session, we initialize these models from pre-trained checkpoints. MobileNetV2 and ResNet18 are pre-trained on the ImageNet-1k~\cite{ILSVRC15}, while DINOv2 is pre-trained on a collection of multiple datasets. For base training, we apply either \textit{(i)} ProtoNet loss~\cite{snell2017prototypical}, yielding \textbf{MobileNetv2-PN}, \textbf{ResNet18-PN}, and \textbf{DINOv2s-PN} pre-trained models, or \textit{(ii)} state-of-the-art methods (\textbf{SAVC}, \textbf{OrCo}, and \textbf{FACT}). 
We select the checkpoint with the best validation accuracy on the base classes after fine-tuning with a slow learning rate.
We also include a non-adapted DINOv2s model, with checkpoint taken from initial contrastive learning pretraining on large vision dataset.
The corresponding backbone is denoted as \textbf{DINOv2s-init}.

\textbf{Evaluation datasets.} 
We choose CUB200~\cite{wah2011caltech}, a common FSCIL fine-grained dataset, CIFAR100~\cite{krizhevsky2009learning} and CORe50~\cite{lomonaco2017core50}, picked specifically to evaluate DINOv2 model on a dataset not seen during self-supervised training\footnote{A full list of datasets used in DINOv2 pretraining is in Table 18 of~\cite{oquabdinov2}.}.
In each dataset, we fix $N_0$ base classes for base training and use the remaining classes as novel classes during incremental sessions. IN CUB200 we set $N_0=100$, in CIFAR100 $N_0=50$, and in CORe50 $N_0=40$.
We conduct 25 evaluation episodes, with each episode involving random subsampling of $N_1$ novel classes, followed by selecting one support sample per novel class.
We then use the chosen novel classes and support samples in few-shot evaluation.

We report results for two ultra-low-data scenarios: $N_1=1$ (one novel class) and $N_1=5$ (five novel classes), focusing on the challenging one-shot setting ($K=1)$, where only one support sample is provided for each novel class.

\subsection{Evaluation Metrics}

Our evaluation metrics are BCR, NCR and FOR from Eqs.~\eqref{eq:bcr}, \eqref{eq:ncr}, \eqref{eq:for}.
We compare accuracy of our NCR (Eq.~\ref{eq:dr}) to the baseline vanilla inference method (Eq.~\ref{eq:vanilla}).

For base class recognition accuracy, we include BCR scores using simple nearest centroid method for base classes.
The BCR metric is the same for both inference methods.

For vanilla inference method, we include NCR metric (denoted by \textit{V-NCR} in the table).
We don't include FOR metric, since it is negligible yet uncontrolled for the frozen backbone during incremental training stage. %

For inference with our NCD, we can select the distance threshold $\alpha$ depending on the bearable \textbf{forgetting budget} for the target application.
For example, for $\alpha=0$, all incoming samples are detected as novel, resulting in high NCR, but complete forgetting of the base classes ($\mathrm{FOR}=\mathrm{BCR}$).
On the other hand, big $\alpha$ level would result in 0\% FOR but also 0\% NCR.

To mimic a practical application where we are willing to trade some BCR for increased NCR, we choose two levels of forgetting budget: $\mathrm{FOR}=2\%$ and $\mathrm{FOR}=5\%$.
We find $\alpha$ values corresponding to those two levels of forgetting, and report two NCR metrics, denoted in the table as \textit{NCR@2FOR} and \textit{NCR@5FOR}, respectively.

\subsection{Main Results and Discussion}

Table~\ref{tab:main} shows NCR comparison between vanilla inference method (\textit{V-NCR}) and inference based on our NCD rule (\textit{NCR@2FOR} and \textit{NCR@5FOR}).
The effectiveness of NCD rule shows some insight into organization and evolution of the feature space for different base training methods.

\textbf{ProtoNet supervised base training.}
As we see from the table, NCR accuracy improves greatly when applied on top of simple ProtoNet base training.
For MobileNetv2-PN and ResNet18-PN backbones with one novel class on CUB200 dataset, NCR is improved by 9.3-10.8\% for the price of 2\% FOR, and by 27-31.3\% for the price of 5\% FOR.
With five novel classes on CUB200 dataset with the same backbones, NCR is improved by 10-16.1\% for the price of 5\% FOR. Similar gains are also achieved by RssNet18-PN on CIFAR100 dataset.

Intuitively, ProtoNet training on base classes with slow learning rate gradually deforms the feature space to cluster the base class samples together.
In the process, the feature space corresponding to novel classes becomes more deformed, so the one-shot prototypes from those classes are more separated from the actual novel class centroid.

As discussed before, we designed the NCD rule to reduce the dependency of the inference result on the choice of the support sample in the novel class.
While \textit{NCR@xFOR} for $N_1=1$ measures pure out-of-distribution capabilities in the feature space (i.e., how well are the base class clusters separated from any novel samples), the same metric for $N_1=5$ also measures
the separability of the novel classes between each other.

The results of MobileNetv2-PN and ResNet18-PN indicate that we gain a lot of accuracy for $N_1=1$, where we rely purely on OOD, but for $N_1=5$ the gains are much smaller since the discrimination ability of those models between unseen classes is worse compared to the contrastive learning-based training methods.

\begin{figure}[tbp]
    \centerline{
        \includegraphics[trim=0cm 12.8cm 15cm 0cm, clip, width=1\linewidth]{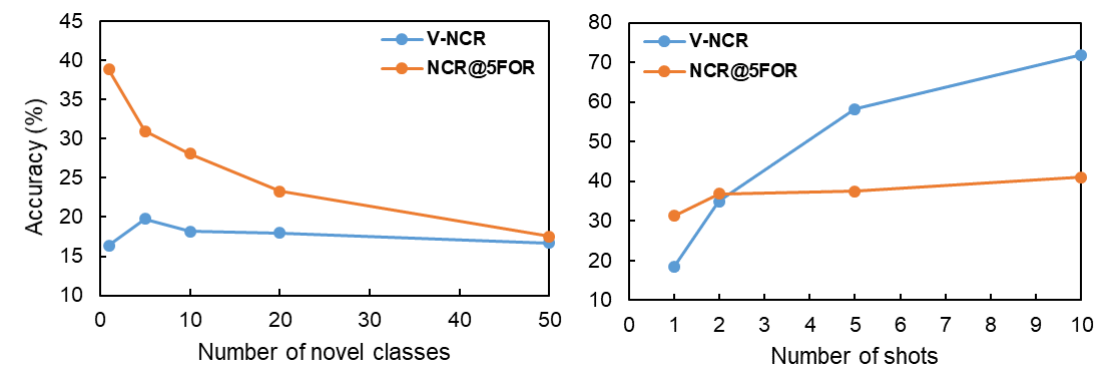}
    }
    \caption{Comparison between vanilla and NCD-based inference methods. Left: NCR for increasing number of novel classes $N_1$ with $K=1$. Right: NCR accuracy for increasing number of shots $K$ with $N_1=5$. Experiments are done on ResNet18-PN backbone with CUB200 dataset.}
    \label{fig:novel_cls_dep}
\end{figure}

\textbf{SOTA base training approaches.}
A better separation of the feature space achieved during pre-training by using SAVC, OrCo, and FACT allows to reach higher results in terms of NCR than ProtoNet-based pre-training.
While vanilla NCR shows good results when applied to SOTA methods, our decision rule can still improve the performance of novel classes, but for a higher FOR price. 

For example, in ResNet18-SAVC trained for FSCIL task, the NCR gains are lower but still notable for CUB200: 22.3\% gain for 1 novel class, and 13.5\% gain for 5 novel classes for the price of 5\% forgetting, with similar gains for MobileNetv2-SAVC backbone.

This suggests superior novel class separation capabilities of backbones trained with contrastive learning (e.g., in SAVC and FACT) and orthogonality promoting terms (e.g., in OrCo).

\textbf{ProtoNet base training on DINOv2 checkpoint.}
Notably, our NCD helps with DINOv2 transformer architecture, with NCR gains on the CORe50 dataset of 18.7\% and 26.2\% for 1 novel class at a price of 2\% and 5\% forgetting, respectively.

Comparing DINOv2-PN training with MobileNetv2-PN and ResNet18-PN, we can see that the initial checkpoint makes a big difference, since DINOv2 was trained via contrastive learning on a large vision dataset, already starting with a highly-separable feature space for classes.
The clustering in DINOv2 feature space is also well-transferable to never-seen-before classes of CORe50, as seen in DINOv2-init metrics.
Starting from that checkpoint, ProtoNet training with small learning rate increases base class recognition rate, but also improves out-of-distribution capabilities (as seen in $N_1=1$ case), as well as keeping the novel classes separated (as seen in $N_1=5$ case).

\subsection{Ablations}

We measure the effectiveness of our NCD rule for varying number of novel classes $N_1$ and number of shots $K$.
As we see from Fig.~\ref{fig:novel_cls_dep}, we can improve NCR accuracy considerably for up to 50 novel classes for one-shot recognition.

However, NCD with 5\% forgetting performs same or even worse than vanilla inference when we increase the number of shots.
In other words, our method targets ultra-low shot regimes.
Vanilla inference mode yields better results when 3 or more shots are available for novel classes.

Finally, we remark that in a practical application: (i) the implementation of the final solution could switch between NCD and vanilla inference modes, depending on the number of samples collected for the novel class; and (ii) the controllable forgetting rate in NCD inference can also be adjusted on device depending on the forgetting strategy.

\section{Conclusion}

In this paper, we explored a novel approach to one-shot class-incremental learning based on novel class detection-based decision rule during inference.
Our method can be applied on top of existing training methods for few-shot recognition, and can give quality-of-service guarantees when applied to on-device personalized applications thanks to its \textbf{controllable forgetting} property.

We evaluated our method against the standard inference method and showed its effectiveness across various backbones and datasets on one-shot recognition task.
Overall, we presented a robust and accurate method for one-shot continual class-incremental learning that can be seamlessly combined with any existing pre-training method.

\bibliographystyle{IEEEtran}
\bibliography{refs.bib}

\end{document}